# Fast and Accurate Power Load Data Completion via Regularization-optimized Low-Rank Factorization

*Yan Xia[1], Hao Feng[1], Hongwei Sun[2], Junjie Wang[3], Qicong Hu[4*]*

*[1] School of Electronic Information and Communication Engineering,*
*Chongqing Aerospace Polytechnic, Chongqing, China*
*[2] Dongfang Electronics Co.,Ltd, Yantai, China*
*[3] China Mobile Cloud Capability Center (Suzhou Center), Suzhou, China*
*[4] College of Computer and Information Science, Southwest University, Chongqing, China*
** a1634688370@email.swu.edu.cn*

## Abstract

Low-rank representation learning has emerged as a powerful tool for recovering missing values in power load data due to its ability to exploit the inherent low-dimensional structures of spatiotemporal measurements. Among various techniques, low-rank factorization models are favoured for their efficiency and interpretability. However, their performance is highly sensitive to the choice of regularization parameters, which are typically fixed or manually tuned, resulting in limited generalization capability or slow convergence in practical scenarios. In this paper, we propose a Regularization-optimized Low-Rank Factorization, which introduces a Proportional-Integral-Derivative controller to adaptively adjust the regularization coefficient. Furthermore, we provide a detailed algorithmic complexity analysis, showing that our method preserves the computational efficiency of stochastic gradient descent while improving adaptivity. Experimental results on real-world power load datasets validate the superiority of our method in both imputation accuracy and training efficiency compared to existing baselines.

## 1 Introduction

In recent years, the reliable monitoring and analysis of power load data has become increasingly important for ensuring the stability and efficiency, and intelligence of modern power systems [1], [2]. Applications such as smart grid operation [3], demand-side management [4], and energy forecasting [5] all rely on high-quality, continuous power consumption data. However, in practice, power load data collected from smart meters or sensor networks is often incomplete due to equipment malfunctions, communication errors, or environmental disruptions. These missing values can significantly impair downstream analytical tasks and decision-making processes, highlighting the urgent need for effective data completion techniques [6]-[9].

Low-rank representation learning has become a powerful paradigm for addressing missing data in spatiotemporal power load matrices [10]-[18]. By leveraging the inherent correlations across time and space, these methods aim to uncover a compact structure that captures the essential patterns of electricity consumption. This representation facilitates robust imputation of missing entries and supports scalable and interpretable modelling [19]-[26].

Among these methods, low-rank factorization (LF) techniques have gained prominence due to their simplicity and efficiency [27]-[33]. LF approaches typically approximate the observed matrix as the product of two low-rank latent factor matrices and optimize a regularized objective function that balances reconstruction accuracy with model complexity. However, a key limitation persists: the regularization coefficient is often fixed throughout training, which restricts the model's ability to adapt to dynamic error patterns and potentially leads to suboptimal performance.

In most existing LF-based methods, the regularization coefficient is statically set via grid search or expert knowledge [34]-[40]. While this may yield acceptable results on small or relatively stable datasets, it becomes problematic in dynamic, heterogeneous power environments. A fixed regularization strength may cause underfitting when the model is too constrained or overfitting when the penalty is too weak, particularly under conditions of varying missing patterns, noise levels, or training progress. Moreover, manual tuning is computationally expensive and lacks generalizability across different power grids or temporal contexts.

To address this issue, we propose a novel Regularization-optimized Low-Rank Factorization framework, abbreviated as λ-opt LF, that introduces a Proportional-Integral-Derivative (PID) controller into the training loop. Rooted in control theory, the PID controller dynamically adjusts the regularization coefficient in response to the real-time prediction error. Specifically, we define the control error as the difference between observed and predicted power load values, and use the proportional, integral, and derivative components to respectively capture the immediate discrepancy, accumulated deviation, and the trend of change. This design enables adaptive and fine-grained control of regularization strength throughout the training process, allowing the model to maintain an optimal balance between fitting accuracy and generalization.

The contributions of this paper are summarized as follows:

- We propose a novel low-rank factorization framework for power data completion, termed λ-opt LF, which employs a PID controller to adaptively regulate the regularization coefficient, improving model generalization and robustness without manual tuning.
- We provide a rigorous analysis of algorithmic complexity, demonstrating that the proposed method maintains favourable computational and memory efficiency compared with conventional LF models.
- We perform extensive experiments on real-world power load datasets, showing that λ-opt LF outperforms baseline methods with fixed regularization in terms of both imputation accuracy and convergence speed.

The remainder of this paper is organized as follows. Section 2 reviews background knowledge on low-rank factorization and PID control. Section 3 presents the proposed λ-opt LF framework and its theoretical analysis. Section 4 details the experimental setup and results. Finally, Section 5 concludes the paper and discusses future research directions.

## 2. Background

*2.1 Low-Rank Factorization*

Low-rank factorization is a foundational technique in both collaborative filtering and time series analysis for power systems [41]-[48]. In the context of electricity load forecasting and modelling, it is often assumed that the temporal load data—i.e., the power consumption recorded at various time points across different grid nodes or equipment—can be effectively captured by a low-dimensional latent structure. This assumption allows the model to uncover shared consumption patterns and temporal correlations across the grid, enabling both noise suppression and robust prediction.

Let $R\in\mathbb{R}^{m\times n}$ be the temporal load matrix, where $m$ denotes the number of time points, and $n$ represents different measurement parameters. Typically, $R$ is sparse or partially observed due to sensor failures or communication delays. The objective of low-rank factorization is to approximate $R$ by the product of two low-rank matrices:

$$R \approx \hat{R} = UV^T, \tag{1}$$

where $U\in\mathbb{R}^{m\times k}$ encodes the latent temporal features, $V\in\mathbb{R}^{n\times k}$ encodes the latent spatial (or load-type) features, and $k$ is the latent dimensionality controlling the approximation rank. Each row of $U$ corresponds to a time-specific latent pattern, while each row of $V$ represents the consumption characteristics of a specific measurement parameter. To learn $U$ and $V$, we typically minimize a regularized loss function over the observed set Ω:

$$\mathcal{L}(U,V) = \sum_{(i,j)\in\Omega} \left( R_{ij} - \langle U_i, V_j \rangle \right)^2 + \lambda \left( \left\| U \right\|_F^2 + \left\| V \right\|_F^2 \right), \tag{2}$$

where $<U_i, V_j>$ is the inner product of the latent vectors of time point $i$ and node $j$, $\|\cdot\|_F$ denotes the Frobenius norm, and $\lambda>0$ is a regularization coefficient. This term penalizes overfitting and ensures generalization by controlling the complexity of the latent representations.

Choosing an appropriate regularization parameter $\lambda$ is critical yet non-trivial. A small value may cause the model to overfit the observed load data, especially under high noise or outlier presence, while a large value may oversmooth the latent structure and miss important variations in load patterns.

Optimization is typically performed using stochastic gradient descent (SGD), alternating least squares (ALS), or related iterative methods [49]-[57]. In this work, we adopt the SGD framework due to its simplicity, scalability, and strong performance on large-scale and sparse time-load matrices [58]-[66]. Moreover, SGD is naturally amenable to dynamic data environments, making it suitable for real-time or online power system monitoring and prediction [75]-[84].

*2.2 PID Controller*

In cybernetics, a PID controller continuously adjusts the control input $u(t)$ based on the current error $e(t)$, the accumulated historical error, and the rate of change of the error. The standard PID control law is given by:

$$u(t) = K_P e(t) + K_I \int_0^T e(\tau) d\tau + K_D \frac{de(t)}{dt}, \tag{3}$$

where:

- $e(t)$ is the control error at time $t$, typically defined as the difference between the desired target and the current system state;
- $K_P$ is the proportional gain, determining the strength of the response to the current error;
- $K_I$ is the integral gain, which accumulates past errors and aims to eliminate steady-state offset;
- $K_D$ is the derivative gain, which predicts future error trends and helps suppress oscillations or overshoots.

The proportional term provides immediate corrective action in proportion to the current error. The integral term accounts for the accumulation of past errors, ensuring long-term accuracy and reducing residual bias. The derivative term anticipates the future behaviour of the system by evaluating the rate of error change, thereby enhancing the stability and responsiveness of the controller.

## 3. Methodology

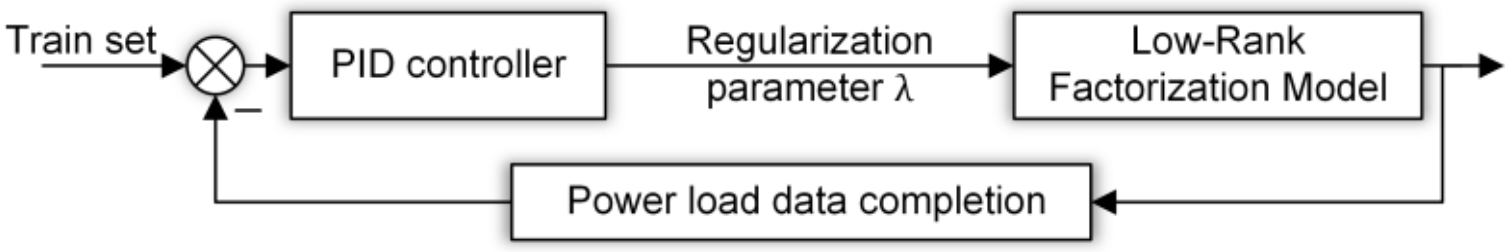

Fig. 1. The flowchart of λ-opt LF framework.

In this section, we present λ-opt, a novel low-rank factorization framework that employs a PID controller to dynamically adjust the strength of regularization terms. Unlike conventional approaches that use a fixed value for the regularization term [67]-[74], our method leverages real-time prediction errors to self-adaptively tune the strength of regularization during training, leading to better balance between model complexity and generalization. Fig. 1. give the flowchart of the λ-opt framework.

### *3.1 λ-opt LF framework*

To adaptively control $\lambda$, we define the control error for each observed entry $(i,j)\in\Omega$ as the prediction residual:

$$e_{ij}(t) = r_{ij} - \langle U_i(t), V_j(t) \rangle. \tag{4}$$

This error reflects how accurately the current model captures the observed load value at time step $t$. We then apply a discrete-time PID controller to adjust the regularization coefficient $\lambda_{ij}(t)$ for each time-load entry, based on the current prediction error, its historical accumulation, and its temporal change:

$$\lambda_{ij}(t) = K_P e_{ij}(t) + K_I \sum_{\tau=1}^{t} e_{ij}(\tau) + K_D \left( e_{ij}(t) - e_{ij}(t-1) \right), \tag{5}$$

where $K_P$, $K_I$ and $K_D$ are positive control gains corresponding to the proportional, integral, and derivative terms, respectively. This PID-based formulation enables the controller to:

- Increase regularization when the prediction error grows or persists over time, helping stabilize training and suppress overfitting;
- Accumulate past errors to respond to long-term bias and slow drift in the data;
- Reduce regularization when the model improves rapidly, allowing more flexible and precise fitting to the observed load data.

To maintain stability and prevent extreme updates, we constrain $\lambda_{ij}(t)$ within a bounded range:

$$\lambda_{ij}(t) \leftarrow \mathrm{clip}(\lambda_{ij}(t), \lambda_{\min}, \lambda_{\max}). \tag{6}$$

Depending on implementation, $\lambda$ can be maintained per time-load pair for computational efficiency.

We integrate the dynamic regularization into an SGD scheme. At each training step:

- Compute prediction error $e_{ij}(t)$, which reflects the discrepancy between the actual and predicted load values for the time-load interaction;
- Update regularization coefficient $\lambda_{ij}(t)$ using the PID rule (5), considering both the current prediction error and its temporal evolution;
- Compute gradients:

$$\begin{cases} g_{U_j} = -2e_{ij}(t)V_j + 2\lambda_{ij}(t)U_i, \\ g_{V_j} = -2e_{ij}(t)U_i + 2\lambda_{ij}(t)V_j. \end{cases} \tag{7}$$

- Perform the updates for the latent matrices $U$ and $V$:

$$\begin{cases} U_i \leftarrow U_i - \eta g_{U_i}, \\ V_j \leftarrow V_j - \eta g_{V_j}. \end{cases} \tag{8}$$

where $\eta$ is the learning rate.

### *3.2 Algorithm and Complexity Analysis*

We provide a detailed analysis of the computational and space complexity of the proposed λ-opt Low-rank Factorization algorithm (Algorithm 1), which uses a PID controller to dynamically adjust the regularization coefficient during sample-wise SGD.

#### *3.2.1 Computational Complexity*

At each iteration, Algorithm 1 performs updates based on a single observed entry $(i,j)\in\Omega$, following the standard pure SGD procedure. In the procedure:

**Line 3: Prediction error computation.** $<U_i,V_j>$ requires $O(k)$ operations, where $k$ is the latent dimensionality.

**Line 4: PID controller update for $\lambda_{ij}(t)$.** This step consists of a few scalar operations and has constant time complexity: $O(1)$.

**Line 5: Gradient computation and updates for $U_i$ and $V_j$.** Each gradient involves two vector operations of size $k$, and the updates are also vector operations. Hence, both

| **Algorithm 1**: $\lambda$-opt Low-rank Factorization | |
|---|---|
| **Input**: Observed entries $\Omega$, learning rate $\eta$, PiD gains $K_P$, $K_I$, $K_D$, $\lambda$ bounds $[\lambda_{\min}, \lambda_{\max}]$ <br> Initialize $U$, $V$ randomly; Initialize $\lambda_{ij}$ for $(i,j) \in \Omega$ | |
| **Output**: $U$, $V$ | |
| 1 | **while not** converge and $t<T$ **do** |
| 2 | **for each** $(i,j) \in \Omega$ |
| 3 | Compute $e_{ij}(t)=r_{ij}-<U_i,V_j>$ |
| 4 | Update $\lambda_{ij}(t)$ with Eq. (5) |
| 5 | Clip $\lambda_{ij}(t)$ to $[\lambda_{\min}, \lambda_{\max}]$ |
| 6 | Update $U_i$ and $V_j$ using SGD with $\lambda_{ij}(t)$ |
| 7 | **end for** |
| 8 | $t = t+1$ |
| 9 | **end while** |
| 10 | **return** $U$, $V$ |

gradient computation and parameter updates have time complexity $O(k)$.

Therefore, total time complexity per iteration in Algorithm 1 is:

$$O(k)+O(1)+O(k)=O(k). \tag{9}$$

Since the algorithm processes each observed entry independently, the total time complexity over one full pass (epoch) over the dataset with $|\Omega|$ observed entries is:

$$O(|\Omega|\cdot k) \tag{10}$$

This matches the time complexity of standard SGD-based low-rank factorization, with negligible additional cost from the PID controller.

*3.2.2 Storage Complexity*

The memory usage of $\lambda$-opt consists of the following components:

**Latent Factor Matrices:** $U\in\mathbb{R}^{m\times k}$ requires $O(m\cdot k)$ space and $V\in\mathbb{R}^{n\times k}$ requires $O(n\cdot k)$ space.

**Regularization Coefficients**: The regularization coefficient vector $\lambda_{ij}(t)$ is stored for each observed entry $(i,j)\in\Omega$. In the worst case, this requires $O(|\Omega|)$ space (if storing per-entry regularization coefficients).

**Error and Gradient Storage**: The prediction errors $e_{ij}(t)$ and gradients are stored for each observed entry. This requires $O(|\Omega|)$ space in total, as each entry requires storage for the error and its gradients (constant space per entry).

Therefore, total space complexity is:

$$O(m\cdot k+n\cdot k+|\Omega|) \tag{11}$$

# 4 Results

*4.1 Experimental Setup*

Table 1 Datasets

| Dataset | Time points | Parameters | Entries |
|---|---|---|---|
| UKDALE [10] | 227,615 | 7 | 309,416 |
| IAWE [11] | 1,081,876 | 13 | 1,557,728 |

**Datasets**. We evaluate our model on two power load datasets, their statistical information is shown in Table I. The hyperparameters $\{\eta, \lambda, K_P, K_I, K_D\}$ on UKDALE and IAWE are empirically set to {5e-2, 9e-4, 5e-2, 5e-4, 5e-4} and {5e-2, 5e-4, 5e-3, 5e-4, 5e-5}, respectively.

**Metrics.** Two commonly used metrics are adopted to evaluate completion accuracy [85]-[94]:

- RMSE (Root Mean Squared Error):

$$\mathrm{RMSE}=\sqrt{\frac{1}{|\Omega_{\text{test}}|}\sum_{(i,j)\in\Omega_{\text{test}}}(r_{ij}-\hat{r}_{ij})^2} \tag{12}$$

- MAE (Mean Absolute Error):

$$\mathrm{MAE}=\frac{1}{|\Omega_{\text{test}}|}\sum_{(i,j)\in\Omega_{\text{test}}}|r_{ij}-\hat{r}_{ij}| \tag{13}$$

**Baseline Methods.** We compare our method against several representative low-rank factorization models, including Momentum-SGD-based LF (MSLF) [12], Nesterov Accelerated Gradient-based LF (NLF) [13], Adam-based LF (ALF) [13], and Nadam-based LF (NALF) [13].

*4.2 Results and Analysis*

Table 2 presents the completion accuracy of our proposed λ-opt LF model compared with four baseline LF models. Fig. 2 shows their convergence time.

As shown in Table 2, λ-opt LF achieves the lowest RMSE (0.5007) and MAE (0.2058) on the UKDALE dataset, outperforming all baseline models. Similarly, for the IAWE dataset, λ-opt LF again attains the best results with an RMSE of 0.2065 and MAE of 0.1053, significantly lower than the baselines. These results indicate that dynamically adjusting the regularization coefficient using the PID-based controller leads to more accurate completion results.

Fig. 2 shows that λ-opt LF converges significantly faster than the baseline models on both datasets. This improvement stems from the adaptive optimization of the regularization parameter, which helps the model avoid prolonged oscillations and accelerates stabilization. Compared to MSLF and NLF, which often require more epochs to reach optimal performance due to fixed regularization, λ-opt LF achieves comparable or better accuracy in fewer iterations.

Overall, the consistent performance gains across both datasets demonstrate the effectiveness of λ-opt LF in capturing temporal patterns in power load data and highlight its advantage over fixed-regularization approaches.

Table 2 Completion Accuracy

| Dataset | Case | λ-opt LF | MSLF | NLF | ALF | NALF |
|---|---|---|---|---|---|---|
| UKDALE | RMSE | **0.5007** | 0.5043 | 0.5084 | 0.5160 | 0.5136 |
| | MAE | **0.2058** | 0.2067 | 0.2094 | 0.2061 | 0.2077 |
| IAWE | RMSE | **0.2065** | 0.2154 | 0.2165 | 0.2279 | 0.2274 |
| | MAE | **0.1053** | 0.1069 | 0.1080 | 0.1124 | 0.1121 |

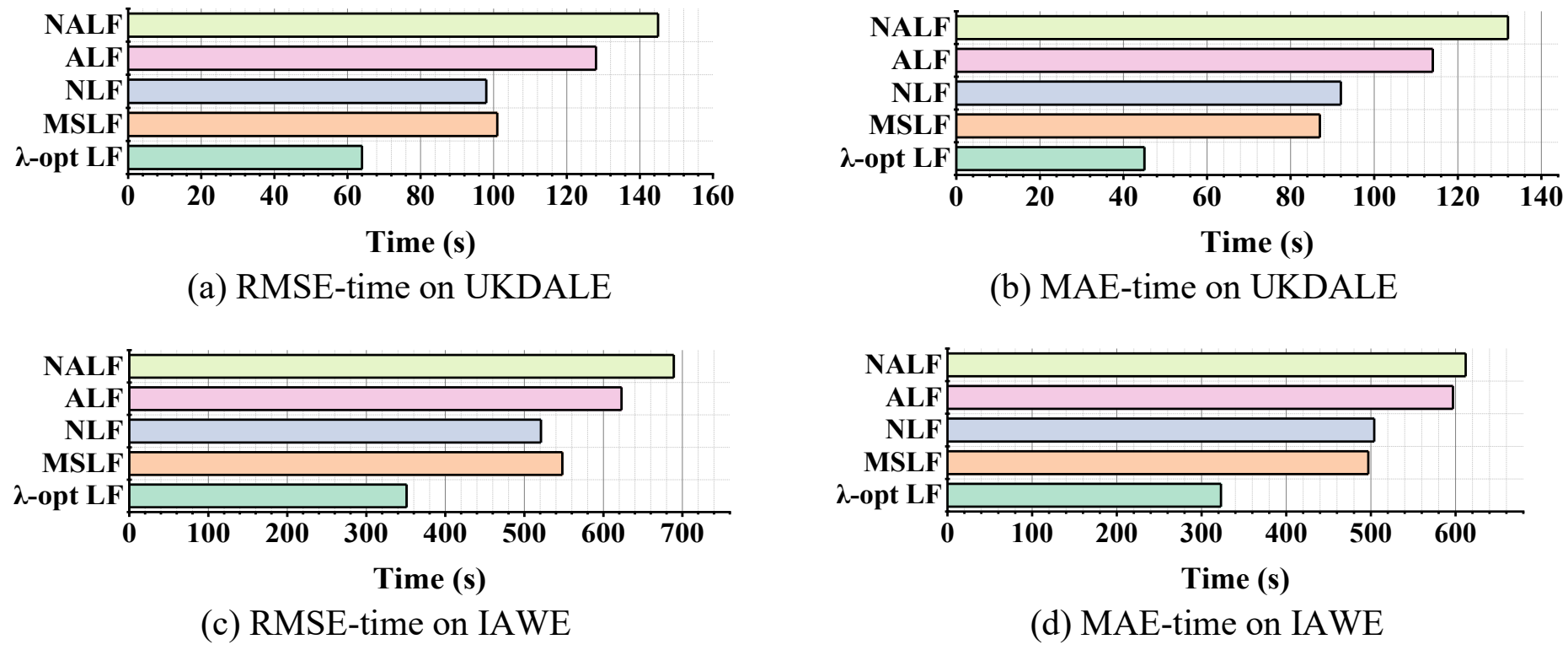

(a) RMSE-time on UKDALE (b) MAE-time on UKDALE

(c) RMSE-time on IAWE (d) MAE-time on IAWE

Fig. 2. Convergence Time

# 5 Conclusion

In this paper, we propose the λ-opt LF framework for power load data completion. Unlike traditional methods with fixed regularization [95]-[98], our approach employs a PID controller to dynamically adjust regularization strength based on real-time prediction errors, achieving a better trade-off between model complexity and generalization. Experiments on two real-world electricity datasets validate its superior accuracy and robustness. Future work will explore advanced adaptive control strategies and extensions to temporal and multi-source load data [99], [100].

# 5 Acknowledgements

This research was supported by the Science and Technology research project of Chongqing Aerospace Polytechnic 2024-XJKJ-04, and the Science and Technology Research Program of Chongqing Municipal Education Commission under grant KJQN202403017, KJQN202403016.